# From Reconstruction to Interpretation: Zero-Setup Multi-Phase Segmentation of X-ray Tomography Data

Pradyumna Elavarthi[1,3], Arun J. Bhattacharjee[2], Harrison Lisabeth[2], Anca Ralescu[1],Petrus H. Zwart[4], Dilworth Parkinson[3] and Elizabeth Clark[3]

[1] Department of Computer Science, University of Cincinnati.
[2] Energy Geosciences Division, Lawrence Berkeley National Laboratory.
[3] Advanced Light Source, Lawrence Berkeley National Laboratory.
[4] Center for Advanced Mathematics in Energy Research Applications, Lawrence Berkeley National Laboratory.

## Abstract

X-ray tomography enables the nondestructive characterization of materials, and recent advances in micro-CT imaging technology have accelerated the acquisition and reconstruction of volumetric data. However, rapid scientific interpretation remains limited by image segmentation, which typically requires manual thresholding, user prompting, or material-specific model training before meaningful analysis can begin. We present a zero-setup deployment framework for multi-phase segmentation of synchrotron X-ray tomography data that produces interpretable segmentations of previously unseen tomography datasets without user prompting or additional retraining during deployment. The framework combines a material-agnostic mask preparation strategy based on common structural concepts in tomography data such as background, sample, bright regions, dark-gray regions, light-gray regions, and porosity, with a pretrained semantic segmentation network to generate diagnostic-quality segmentations within minutes of reconstruction. Unlike conventional deep learning pipelines that require dataset-specific annotations and retraining, the proposed approach can be applied immediately to new scans, enabling rapid assessment of experimental quality and preliminary microstructural analysis during ongoing beamline experiments. Users can subsequently refine or fine-tune the generated masks when application-specific segmentation is required. Experiments on held-out and qualitatively evaluated synchrotron micro-CT datasets suggest that the framework produces consistent and physically meaningful segmentations on new tomography datasets while substantially outperforming conventional intensity-based thresholding. By bridging the gap between high-speed data acquisition and immediate interpretation, the proposed framework enables near real-time beamline feedback and provides a practical foundation for scalable AI-assisted scientific imaging workflows.

# 1. Introduction

X-ray tomography is a non-destructive imaging technique widely used for studying the 3D microstructure of materials across diverse scientific disciplines such as geosciences, material science, biosciences and archaeology [1], [2], [3]. Using X-ray tomography, the 3D morphology and distribution of different phases, regions of interest and defects such as pores, cracks and fractures can be analyzed [4]. The high flux and coherent beam of synchrotron-based X-ray tomographic imaging have the added advantage of shorter data acquisition time and better phase contrast in samples compared to lab-based tomography techniques. Environmental cells can be used to create desired pressure and temperature environments around samples at synchrotron tomography beamlines [5]. This allows analysis of the evolution of the sample microstructure with time.

X-ray tomography beamlines at synchrotron facilities around the world attract hundreds of users per operating cycle and can produce terabytes of volumetric data in a day [6]. Despite the advances in resolution and speed of reconstruction through optimized reconstruction pipelines, downstream data analysis has been a substantial bottleneck [7]. Often, a complete analysis of the data using manual or semi-automated segmentation of volumetric data can take up to several months due to the combination of data management tasks, data processing tasks, the need for painstaking manual interaction, and the lack of sufficient computational resources to parallelize analysis. It is further exacerbated when morphology is inconsistent across the volume. Segmentation models may also fail to generalize under varying experimental conditions, such as uneven illumination, reconstruction artifacts, noise, and changes in beam or scan parameters.

Furthermore, in-situ experiments track the evolution of microstructures during different processes such as heating, mechanical testing and solidification. Complete or partial failure of experiments over multiple temporal measurements are not always evident during the experiment and come to light after completion of careful and dedicated analysis of the data typically weeks to months after the experiment was conducted. Addressing this issue requires real-time or near real-time analysis to help decide whether the experiment can continue, be repeated or requires changing parameters. Near real-time data analysis could save valuable beamtime that would otherwise be spent collecting unusable data.

In this work, we introduce a zero-setup deployment framework for multi-phase semantic segmentation of synchrotron tomography data. The framework can be applied immediately to previously unseen reconstructed datasets without user prompting, material-specific annotations, or additional retraining. Unlike interactive segmentation methods that require user-provided prompts or iterative refinement, the proposed framework automatically generates a complete set of interpretable semantic masks immediately after image reconstruction that enable rapid inspection of sample morphology, porosity, density variations, and other structural features. This allows beamline users to evaluate experiments while they are still running, identify failed scans, and decide whether to continue, repeat, or modify acquisition parameters.

The primary objective of the proposed framework is not to replace application-specific segmentation models, but to provide an immediate, zero-setup segmentation that enables rapid beamline diagnostics and preliminary scientific analysis. The generated masks provide an interpretable first-pass segmentation that allows researchers to quickly assess sample quality, identify regions of interest, and evaluate experimental outcomes while data acquisition is still in progress. When higher segmentation accuracy or material-specific labeling is required, these predictions can be manually refined or used as the starting point for fine-tuning application-specific models.

This capability is enabled by a material-agnostic mask preparation strategy, multi-label semantic segmentation, class-aware sampling, a lightweight ConvNeXt-UNet based architecture[8], class-balanced multi-label learning [9], domain-robust preprocessing strategies, and adaptive sampling methods tailored to scientific tomography [10]. Together, these components enable the framework to produce rapid, diagnostic-level segmentations of previously unseen tomography datasets without additional retraining. The proposed framework was evaluated on datasets from three rock cores exhibiting substantial differences in phase contrast and microstructural texture, demonstrating its applicability across diverse tomography data.

Rather than pursuing highly specialized, material-specific segmentation, the proposed framework is intentionally designed to provide rapid, diagnostic-level segmentation immediately after image reconstruction. By representing tomography data using a small set of broadly occurring and physically interpretable semantic regions such as background, sample, bright regions, dark-gray regions, light-gray regions, and porosity, the framework captures structural concepts that are common across a wide range of multi-phase tomography datasets. This enables researchers to obtain meaningful segmentations of previously unseen samples with zero user setup, facilitating rapid beamline diagnostics and preliminary scientific analysis. The framework is not intended to replace application-specific segmentation models; instead, it provides an interpretable starting point that can be refined manually or used to fine-tune specialized models when higher accuracy or material-specific labeling is required. In this way, the proposed approach bridges the gap between near real-time tomography reconstruction and actionable scientific interpretation during synchrotron experiments.

### 1.2. Image Segmentation for X-ray tomography

Image segmentation is the process of assigning a class label to individual pixels in an image such that regions belonging to the same structural or material category can be isolated, analyzed, and quantified independently. In the context of X-ray tomography, segmentation enables the separation of distinct material phases and defects (such as pores and cracks within rocks). These class assignments can then form the basis for qualitative observations and quantitative measurements such as volume fraction and morphology. Traditionally, this task has been carried out using image analysis software tools where the user manually or semi-automatically defines pixel intensity thresholds and crops regions of interest for the volumetric data. Such approaches rely on the pixel intensity and contrast levels being uniform across the entire image stack which in practice is difficult to achieve due to subtle fluctuations of the X-ray source.

To address the problem of non-uniform pixel intensity and phase contrast across the image stack, advanced segmentation methods such as global and local thresholding [11], Otsu's method [12], region growing [13], [14], edge detection [15], watershed transforms [16], and morphological operations such as erosion and dilation [11], [15] were developed. These methods assume that material phases can be separated primarily by their intensity values or local gradients. While such techniques can perform well in scenarios with high contrast and clearly separable phases [17], they often break down in real tomography data where intensity distributions of different phases or separate regions of interest often overlap, reconstruction and noise artifacts are present, and boundaries between phases are ambiguous. In synchrotron-based tomograph imaging datasets, differences in beam energy, sample composition and scattering effects further complicate the segmentation task, making intensity-based criteria alone insufficient to consistently distinguish between meaningful structural regions [18].

Recent advances in machine learning have therefore enabled a paradigm shift in how segmentation may be performed [19], [20]. In this process, instead of manually or heuristically defining rules for segmentation, a small representative set of images can be segmented using conventional tools such as ImageJ [21] or Dragonfly [22] to generate ground-truth masks with high levels of accuracy. These labeled images can then be used to train a neural network to automatically learn the visual, textural, and structural patterns associated with each region of interest. Once trained, the model can rapidly and consistently segment new images from the same or even different experiments involving similar samples [23]. This data-driven approach significantly reduces human effort and enables real-time or near-real-time data interpretation, making it particularly well-suited for high-throughput synchrotron-based micro-CT experiments and in-situ studies. In this paper, we follow this data-driven approach to prepare masks representing widely occurring structural concepts in tomography data and use them to train a deep neural network to segment a wide variety of tomographic images into intuitive masks without the need for retraining.

# 2. Methodology

## 2.1 Data acquisition

The images used for training our models were selected from multiple scans at X-ray micro tomography beamline 8.3.2 at the Advanced Light Source, Lawrence Berkeley National Laboratory. The annotated dataset consisted of 25 reconstructed 2D slices selected from five tomography scans/material systems. Twenty slices were used for training and five additional slices were held out for evaluation. The distribution of annotated slices across materials and splits is summarized in Table 1.

Table 1. Summary of annotated tomography slices used for model training and evaluation. The dataset contains 25 manually annotated reconstructed slices from five tomography scans. For scans with a train/test split, the values in the third column indicate the number of training and testing slices, respectively.

| Material / scan name | Number of annotated slices | Split | Resolution (pixels) |
|---|---|---|---|
| Aluminum Alloy 1 | 4/1 | Train/Test | (2560, 2560) |
| Duprow (Dolomite) | 4/1 | Train/Test | (2560, 2560) |
| Aluminum Alloy 2 | 4/1 | Train/Test | (2560, 2560) |
| Poorman ( Schist ) | 4/1 | Train/Test | (2560, 2560) |
| Basalt | 4/1 | Train/Test | (2560, 2560) |
| **Total** | **25** | **20 train / 5 test** | |

## 2.2 Mask preparation

To train the segmentation model, we constructed a curated set of semantic masks derived from reconstructed tomographic slices obtained at ALS beamline 8.3.2. Twenty-five representative image slices were selected from multiple geological and metallurgical classes, including serpentinite rock and aluminum alloy exhibiting diverse features. These samples were chosen to reflect the broad variability commonly observed in synchrotron tomography datasets, ensuring that the learned model would not be overfit to any specific material type.

Ground-truth masks were prepared using the commercial segmentation software Dragonfly ORS [22], which provides advanced tools for tomographic image analysis. The segmentation protocol in Dragonfly involved initial intensity-based thresholding for each of the different classes followed by refining operations based on largest connected component. Voxel filling operations were used for pixels that the intensity based segmentation failed to pick up inside the regions for each of the classes. The ‘sample’ mask was prepared by adding dark gray, light gray, bright and porosity masks and the background mask is its inverse. The segmentation protocol explicitly targeted six semantic regions: background, sample, bright, light gray, dark gray and porosity, which are widely present across tomographic datasets. An example of the classification is shown in Fig. 1 for a typical reconstructed image shown in Fig. 1 (a). Details of these mask classes are as follows:

- **Mask 1**: Mask generated by combining all material phases including porosity (except '*background*'). This represents the entire volume of the sample. We call this mask '*sample*' as shown in Fig. 1(b).

- **Mask 2**: Pixels corresponding to empty space or regions surrounding the sample that are not of interest for analysis of the data. Background identification ensures that sample-free regions do not interfere with downstream intensity-based segmentation. We call this mask '*background*' as shown in Fig. 1 (c).
- **Mask 3**: Lowest intensity regions were grouped based on their darker-gray intensity distribution. We call this mask '*dark gray*' as shown in Fig. 1 (d).
- **Mask 4**: Intermediate intensity regions within the sample were segmented as a separate class. We call this mask '*light gray*' as shown in Fig. 1 (e).
- **Mask 5**: Highest intensity regions that typically appear as regions with the brightest pixel intensity. We call this mask '*bright*' as shown in Fig. 1 (f).
- **Mask 6**: Pores or cracks inside the sample commonly share the intensity with the background. These regions have the lowest pixel intensity. We call this mask '*porosity*' as shown in Fig. 1 (g).

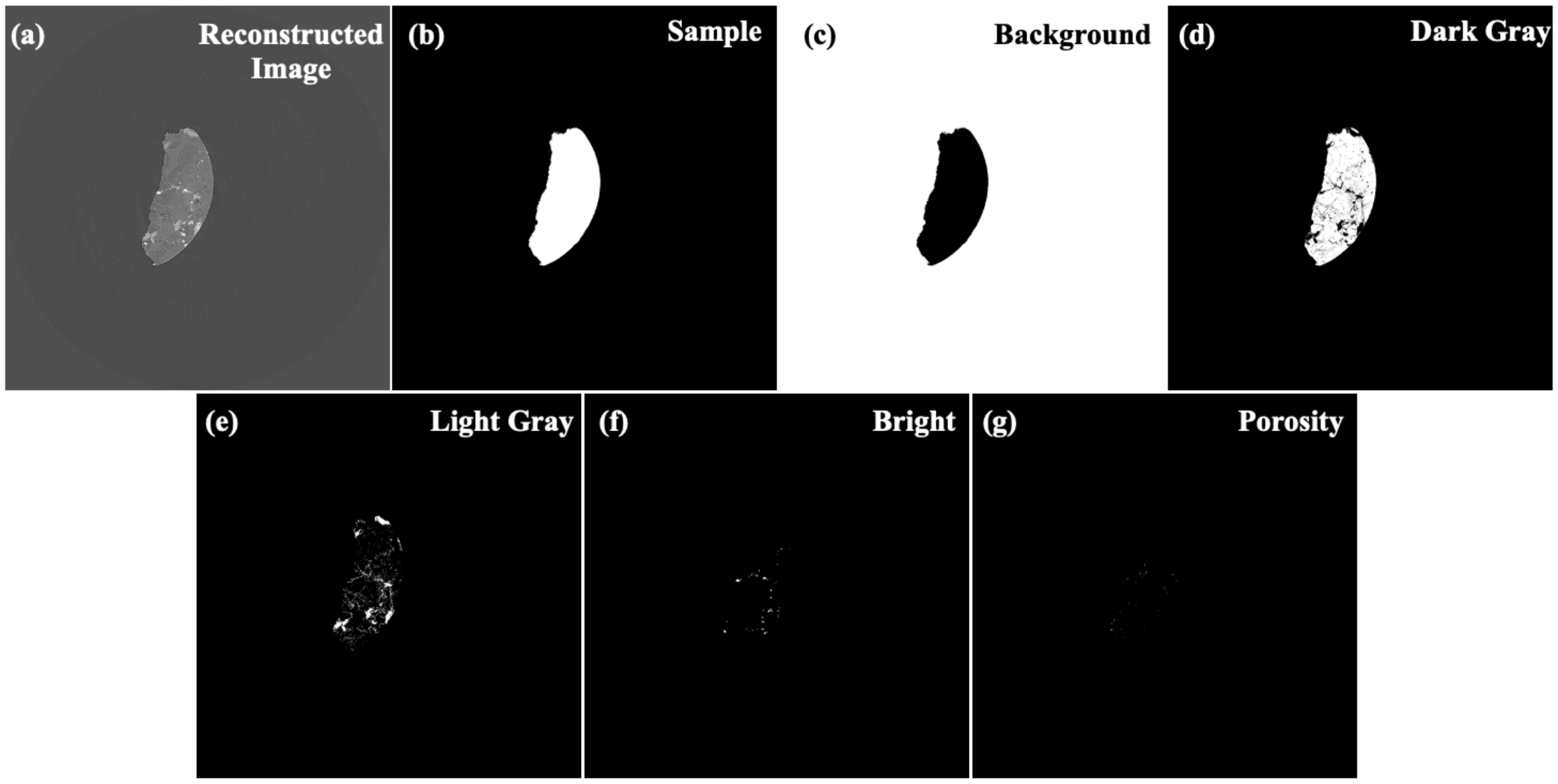

Figure 1: Material-agnostic mask preparation strategy for multi-phase micro-CT segmentation. (a) Reconstructed image. (b) Sample mask. (c) Background mask. (d) Dark-gray region mask. (e) Light-gray region mask. (f) Bright region mask. (g) Porosity mask.

Each region was extracted as a binary mask, and the six channels were stacked to form a multi-label semantic tensor of shape (6 × H × W) for each slice. Rather than relying on a novel segmentation architecture, the proposed framework derives its main contribution from this semantic representation of tomography data. By decomposing reconstructed slices into a small set of broadly occurring structural concepts, a single pretrained model can produce useful diagnostic-level segmentations immediately after image reconstruction without requiring material-specific label definitions or additional retraining. Unlike conventional semantic segmentation pipelines that enforce strict single-class labeling [24], [25], [26], our approach deliberately decomposes the sample into six physically interpretable and widely occurring categories. This design offers several practical advantages for deployment:

- **Material-agnostic representation:** The chosen mask categories are morphologically and visually present across a large variety of geological, metallurgical, and materials science specimens. Even when the exact chemical composition differs, the absorption-based contrast patterns (bright, light gray, dark gray) remain consistent across material modalities. In the case of samples where one or more of these classes are missing, the model simply predicts a blank mask as no pixels belonging to those classes were detected in the image.
- **Robustness to prediction uncertainty:** When deployed on previously unseen samples, including data acquired from different beamlines or under varying imaging conditions, the model may occasionally confuse visually similar categories (e.g., light gray and dark gray). However, because each mask independently represents a distinct structural or density-related feature, at least a subset of the predicted masks typically remains informative even when some predictions are incorrect. This property is particularly valuable during deployment because useful structural information is often preserved even when some semantic regions are predicted less accurately, allowing the framework to remain informative across tomography datasets acquired under different imaging conditions.
- **Utility for downstream scientific analysis:** Porosity, bright spots, and density variations can be direct indicators of different chemical processes. As a result, predictions can provide domain scientists with actionable, physically interpretable insights without any additional training or fine-tuning.

## 2.3. Class aware mapper

In order to deal with the significant class imbalance inherent in micro-CT data, where the vast majority of pixels correspond to background regions, we employ a class-aware cropping strategy during training (Fig. 2). In raw reconstructed tomographic slices, informative structures such as bright regions, light and dark gray regions, porosity, and sample boundaries typically occupy only a small fraction of the field of view. Sampling random crops therefore results in training patches dominated by background, providing little to no usable signal for learning minority structures[27].

To address this, we use a class-aware mapper that selectively extracts training crops (1024,1024) containing at least one non-background class. The mapper attempts multiple (up to 10) random crop locations within an image and retains the first crop where any of the foreground class masks (bright, light gray, dark gray, porosity, sample) contain non-zero pixels. If no such crop is found after several attempts, a centered fallback crop is used to maintain spatial diversity. This strategy ensures that each minibatch contains semantically meaningful and informative regions, substantially increasing the frequency with which minority classes contribute to gradient updates.

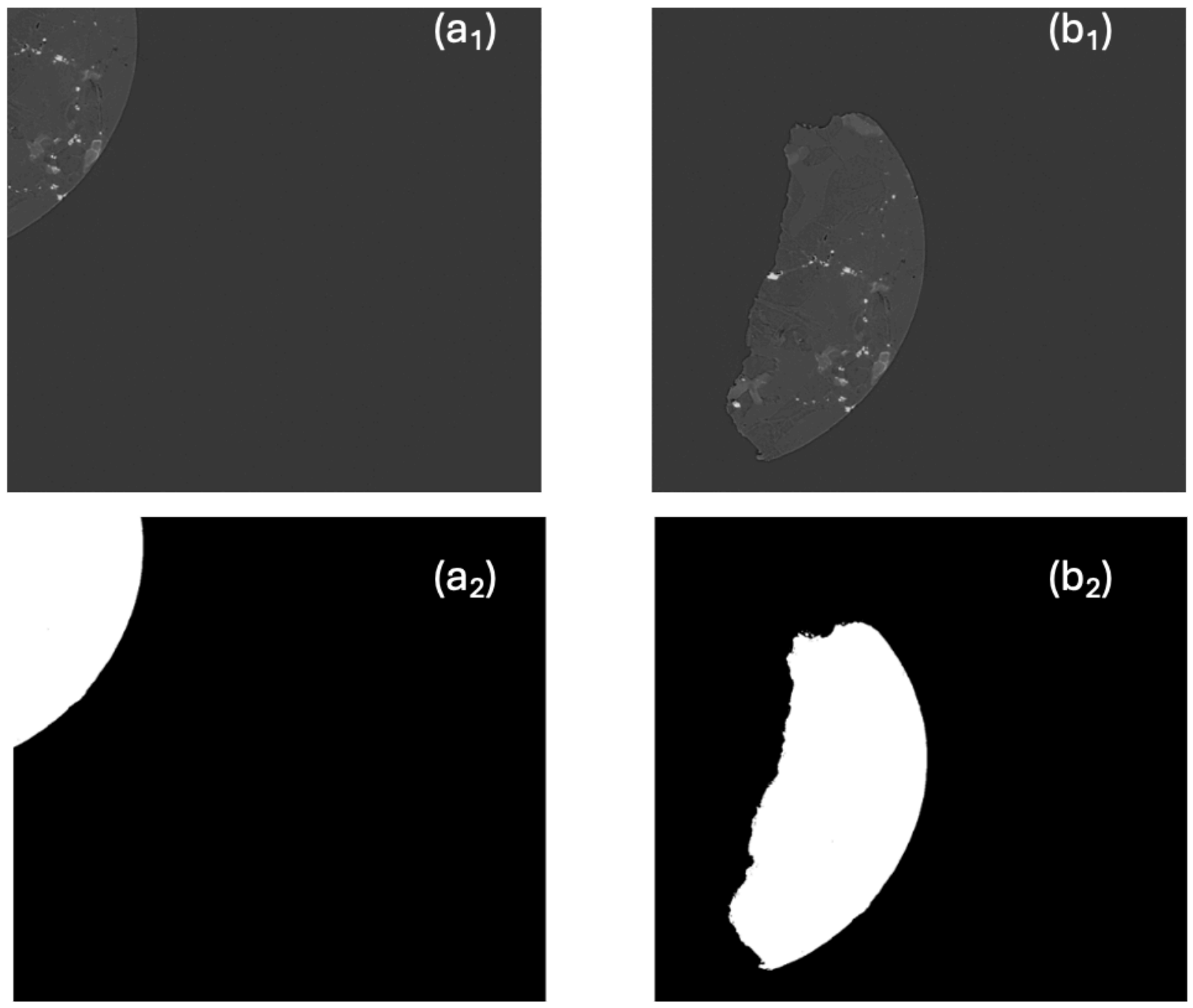

Figure 2: Class aware sampling strategy. ($a_1$–$a_2$) Data sampled randomly. ($b_1$–$b_2$) Data sampled using the class aware mapper.

As shown in Fig. 2, a purely random cropping strategy frequently selects patches dominated by background($a_1$-$a_2$), whereas the class-aware mapper ($b_1$-$b_2$) prioritizes regions containing actual material structure. This targeted sampling significantly enhances the diversity and informativeness of the training data presented to the network, improving the learning of minority structures while reducing the dominance of background-only patches. This contributes to more stable segmentation performance across tomography datasets exhibiting different microstructures and contrast characteristics.

## 2.4 Model

To perform a multi-label segmentation, we developed a high performance segmentation framework based on the ConvNeXt-UNet architecture, which combines a symmetric U-Net [19] topology with a modern ConvNeXt-Tiny [28] backbone. The model features a 1 x 1 convolutional multi-label classification head and is optimized for multi-label segmentation of heterogeneous attenuation-based structural regions. The architecture of the model is shown in Fig 3.

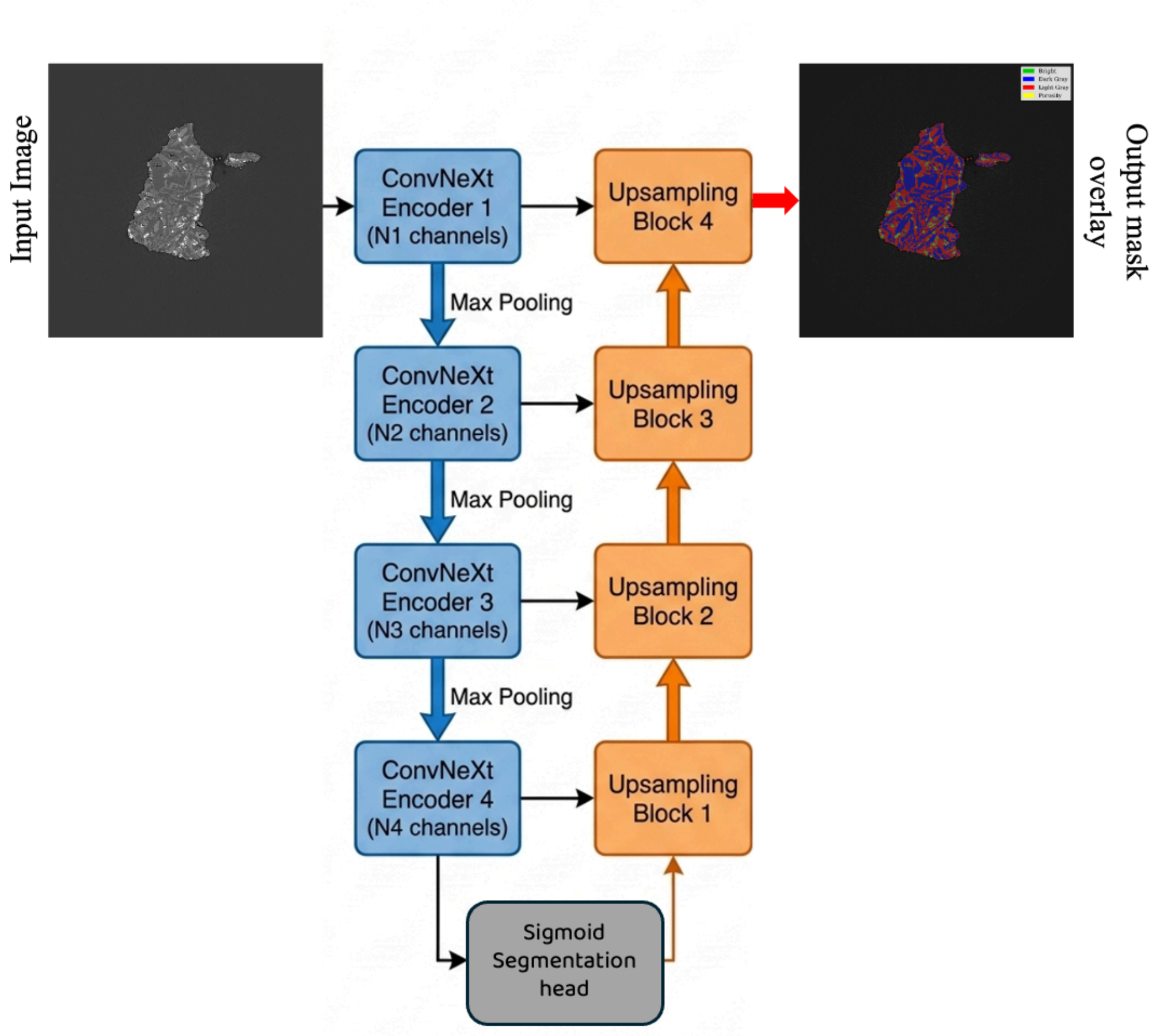

Figure 3: Model architecture used for fine-tuning and inference on tomography images, employing a ConvNeXt-UNet architecture.

### 2.4.1 Input normalization and batching

Reconstructed slices (32-bit grayscale TIFFs) were normalized via percentile-based scaling (1–99%). To make the model invariant to varying contrast conditions across different synchrotron beamlines, we implement a dynamic percentile jittering strategy during training, where the normalization bounds are randomly sampled between 0.01 and 1.5 percentiles ($p_{low} \sim U(0.01, 1.5)$, $p_{high} \sim U(98.5, 99.99)$). Normalized slices are converted to 3-channel RGB tensors to leverage pretrained vision priors. Images are processed in 1024x1024 patches, ensuring that high-frequency textures and phase boundaries are preserved without downsampling artifacts.

### 2.4.2 Network Architecture

We adopt a ConvNeXt-UNet architecture, which replaces standard residual blocks with ConvNeXt blocks. ConvNeXt is a 'purely convolutional' architecture that re-implements the architectural innovations of vision transformers (ViT), such as large kernel sizes (7x7), Layer Normalization instead of Batch Normalization, and inverted bottlenecks, while preserving the inductive biases and efficiency of convolutional neural networks[28]. This hybrid is particularly suited for synchrotron tomography for several reasons:

- **Global-Local Feature Integration:** By using large (7 x 7) kernels in the ConvNeXt blocks, the network achieves a larger effective receptive field than standard ResNets [29], [30], allowing it to capture the long-range structural context (e.g., large mineral grains) while maintaining the pixel-level precision needed for fine cracks.
- **Symmetric Multi-scale Decoder:** Unlike the top-down pathway of a Feature Pyramid Network[31], the U-Net symmetric decoder uses dense skip connections to fuse low-level spatial features from the encoder directly with high-level semantic features. This ensures that the boundaries of minority phases (e.g., bright inclusions) remain sharp and well-localized.
- **Modern Vision Priors:** ConvNeXt blocks utilize a Gaussian Error Linear Unit (GELU)[32] activation and fewer normalization layers, which has been shown to improve convergence stability and generalization on scientific datasets compared to older ResNet-based architectures.

Instead of predicting a single categorical label per pixel, the 1x1 convolutional prediction head outputs six binary channels corresponding to the classes: background, bright, dark gray, light gray, porosity and sample. Multi-label segmentation is preferred because tomography data often contains overlapping material regions, multi-phase regions, and ambiguous boundaries, where single-label argmax segmentation is overly restrictive. This approach ensures that even when the model is uncertain or misclassifies one class, other predicted masks remain meaningful, providing interpretable intermediate layers for scientific analysis and downstream quantitative pipelines[33].

### 2.4.3 Training

The ConvNeXt-UNet model is initialized with ImageNet-1K pretrained weights for the ConvNeXt backbone, ensuring that the network begins with robust feature detectors for edges, textures, and contrast gradients. The model is trained using Binary Cross-Entropy (BCE) with Logits loss. To address the severe class imbalance inherent in tomography data, we apply Median Frequency Balancing[10], which assigns higher weights to rare classes such as porosity and bright inclusions. Training was performed on an NVIDIA A100 GPU for 15,000 iterations. The loss curve is shown in Fig. 4

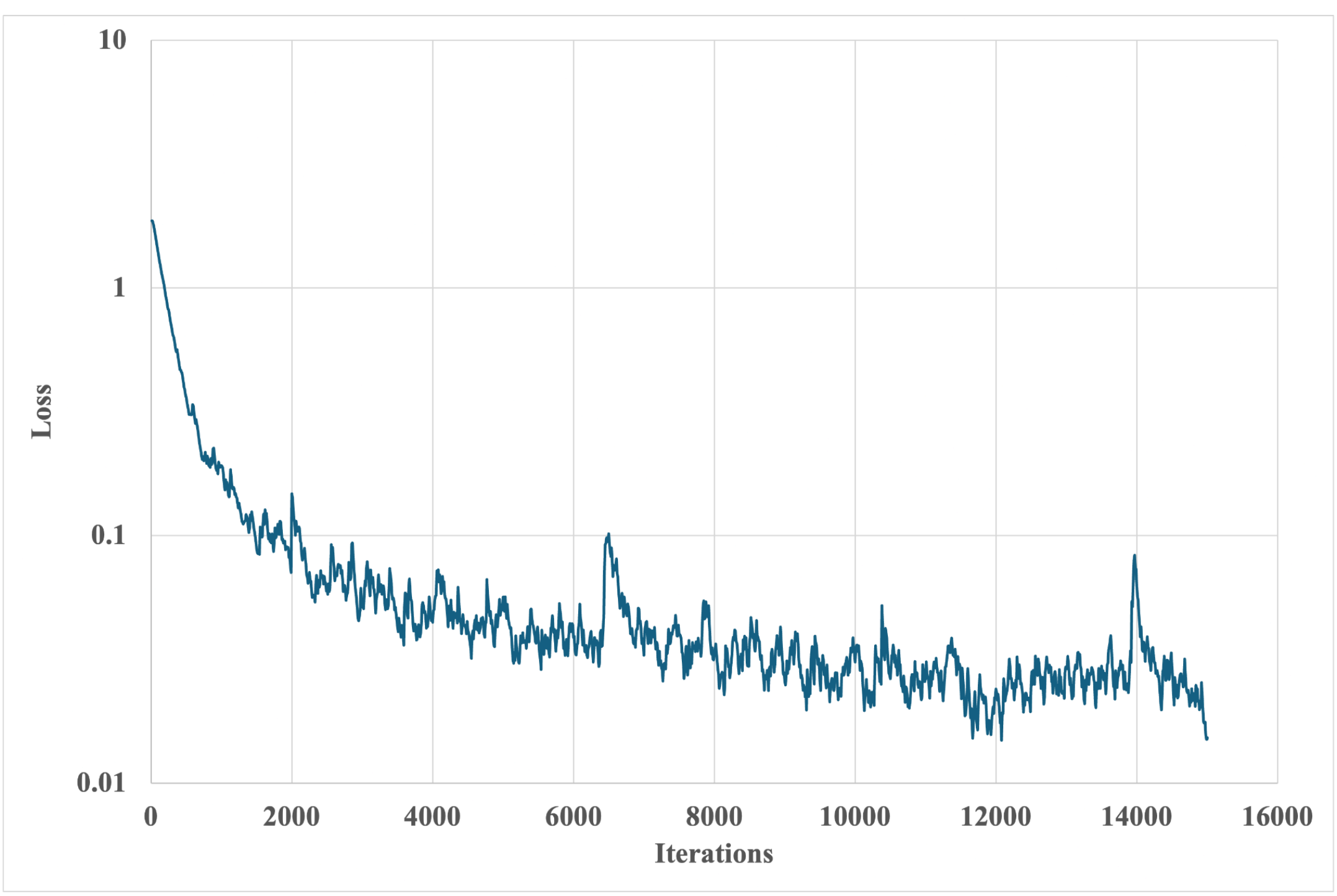

Figure 4: Evolution of training loss till 15000 iterations

### 2.4.4 Per-class weighting and Median frequency balancing

To address the severe class imbalance inherent in tomographic datasets, where background pixels dominate and minority structures (e.g., porosity or bright inclusions) comprise only a small fraction of the image,we incorporate per-class weighting based on the empirical pixel frequency of each mask channel across the training set [5], [26]. For each class $c$, we compute the number of foreground pixels $p_c$ and obtain its relative frequency:

$$f_c = \frac{p_c}{\sum_k p_k}. \quad (1)$$

Here,

- $p_c$ denotes the total count of pixels belonging to class $c$ across all training images,
- $\sum_k p_k$ is the total number of labeled pixels across all classes, and
- $f_c$ quantifies how common a class is relative to the entire dataset.

Rare classes exhibit extremely small $f_c$, which would yield negligible gradients during training unless explicitly compensated [27]. To correct this imbalance, we adopt median-frequency balancing [9], a widely used weighting strategy for semantic segmentation. The weight assigned to each class is computed as:

$$w_c = \frac{median\{f_k | f_k > 0\}}{f_c + \epsilon}, \tag{2}$$

Here,

- $median\{f_k\}$ is the median of all non-zero class frequencies, used as a stable reference baseline,
- $f_c$ is the relative frequency of class $c$, and
- $\epsilon$ is a small constant added to avoid division by zero.

Intuitively, if a class appears less frequently than the median class, it receives a larger weight $w_c$, amplifying its contribution to the loss. Conversely, very common classes receive smaller weights. Incorporating these weights into the BCEWithLogits loss ensures that rare but scientifically important structures such as pores, cracks influence during optimization, preventing the model from collapsing to background-dominant predictions.

# 3. Results and Discussion

To evaluate the performance of the proposed segmentation framework, we computed the F1 scores [34] across the test set for different architectures. The proposed ConvNeXt-UNet model achieved the strongest overall performance, with an accuracy of 0.995 and a macro F1 score of 0.992, indicating highly consistent segmentation across the six semantic mask categories despite the limited size of the training dataset. These results show that the proposed material-agnostic mask representation, combined with class-aware sampling and domain-robust preprocessing, can support accurate diagnostic-level segmentation of tomography images using only a small curated set of annotated slices..

Qualitative examples of the predicted multi-label masks for schist, dolomite, and basalt rock samples are shown in Fig. 5. For each sample, a representative reconstructed slice is displayed together with the corresponding mask overlay. In the schist sample, the dark-gray and light-gray regions may correspond to mineral phases such as chlorite and quartz, with 3D mask-derived volume fractions of 61.34% and 36.08%, respectively [35]. The dolomite sample is dominated by a single attenuation region consistent with the expected dolomite-rich composition along with calcite [36]. In the basalt sample, the light-gray and dark-gray regions may correspond to feldspar- and augite-rich phases, with 3D mask-derived volume fractions of 44.78% and 48.15%, respectively, while the bright region, possibly associated with magnetite-rich inclusions, accounts for 3.49% [37]. These volume fractions were calculated from the predicted region-specific masks over the full 3D volume and normalized by the whole-sample mask. Definitive mineral-phase identification would require complementary characterization, such as X-ray diffraction or chemical analysis, and is beyond the scope of this work.

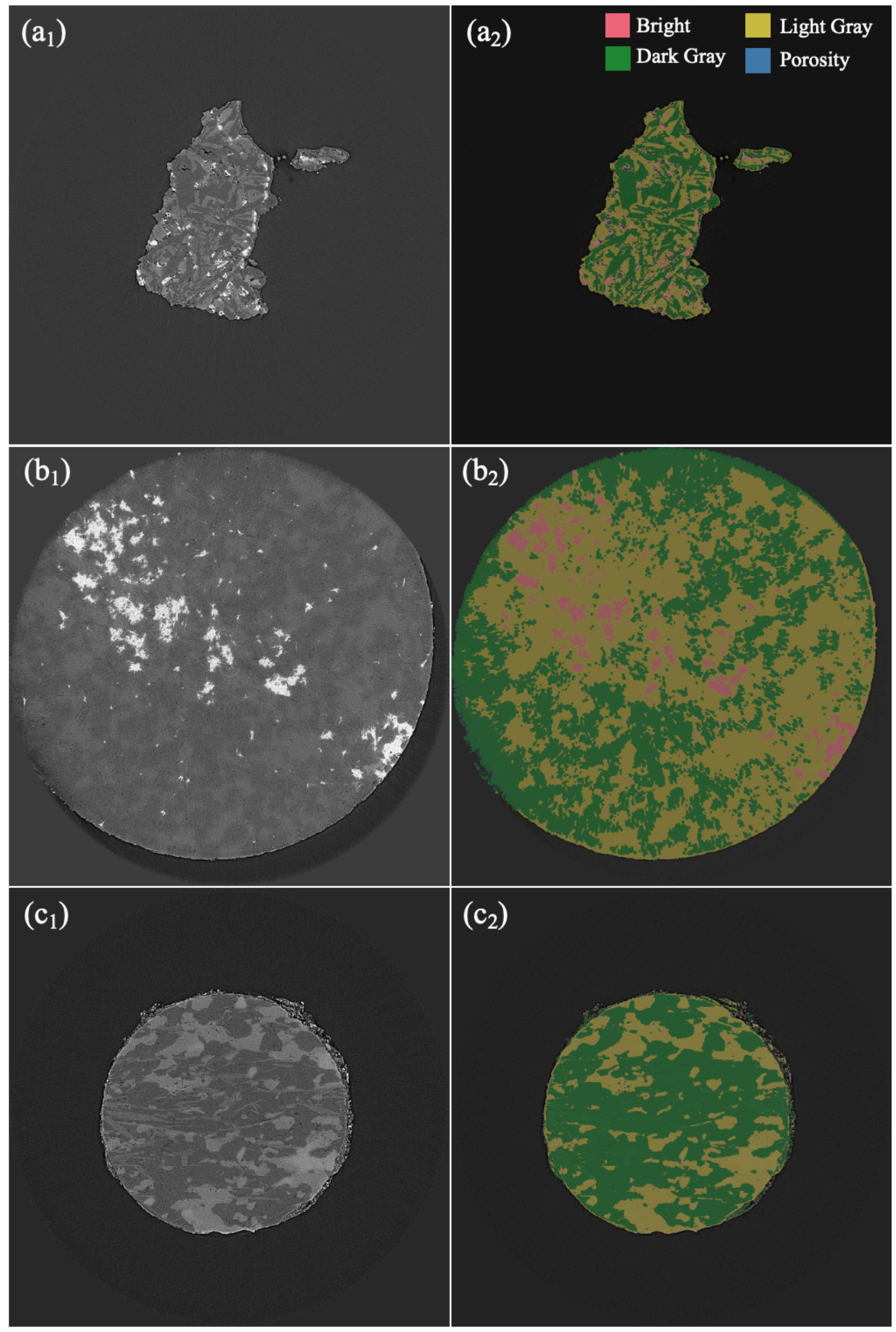

Figure 5: Predicted segmentation masks for three tomography samples: Basalt (a), (b) Dolomite and (c) Schist. For each sample ($a_1$, $b_1$, $c_1$) and ($a_2$, $b_2$, $c_2$) show the predictions overlaid on the original images. Bright, Dark Gray, Light Gray and Porosity are represented in the overlay as Tol Red, Tol Green, Tol Yellow and Tol Blue respectively.

The visual results are consistent with the quantitative metrics by showing that major structural regions are delineated accurately, and fine-scale regions are consistently identified. An important characteristic of the framework is its ability to generalize across previously unseen samples without retraining. The test data included materials with significantly different morphology, texture, attenuation behavior, and contrast distributions from those present in the training set. Despite this variability, the model consistently generated intuitive segmentation masks that preserved meaningful structural information. However, the model assumes that the sample contains four or less classes of density-based phases including porosity. In the case that the sample contains more than four classes, regions belonging to additional classes will be included in one of the three classes as additional classes generally are not easily distinguishable due to low image contrast from the three classes of the model. The model has been applied on micro-CT datasets of different samples that are scanned at 8.3.2 under varying conditions. The original images and corresponding mask overlays are shown in Supplementary Fig. S1. These results suggest that the proposed mask preparation strategy encourages the network to learn structural representations that remain useful across diverse tomography datasets.

To evaluate the influence of network architecture on segmentation performance, we compared three different backbone models as reported in Fig. 6: a ResNeXt-FPN architecture, a DINOv2-based segmentation framework, and the proposed ConvNeXt-UNet architecture built on a ConvNeXt backbone. All models were trained using the same mask preparation strategy, preprocessing pipeline, and class-aware sampling framework to ensure a fair comparison.

Among the evaluated architectures, the proposed ConvNeXt-UNet framework achieved the best overall performance, obtaining an accuracy of 0.995 and a macro F1 score of 0.992. The strong performance of ConvNeXt-UNet can be attributed to the combination of ConvNeXt blocks and the symmetric U-shaped decoder structure, which effectively preserves both global contextual information and fine-scale spatial details. The architecture demonstrated particularly strong boundary preservation and stable segmentation of complex morphological regions.

The ResNeXt-FPN[30] model also produced competitive results, achieving an accuracy of 0.995 and a macro F1 score of 0.991. This indicates that conventional CNN-based architectures remain highly effective for tomography segmentation tasks when combined with the proposed mask preparation strategy. The feature pyramid structure enabled effective multi-scale feature extraction, especially for large structural regions and sample boundaries.

The DINOv2-EoMT[38] based model achieved slightly lower performance, with an accuracy of 0.986 and a macro F1 score of 0.978. Although DINOv2[39] provides strong semantic representations learned through self-supervised learning, the transformer-based features were comparatively less effective at preserving extremely fine local structural details and sharp material boundaries in tomography images perhaps due to losing global context while patching and small dataset sizes. Nevertheless, the model still demonstrated strong overall segmentation capability and generalized well across unseen samples, highlighting the potential of foundation vision models for scientific imaging applications.

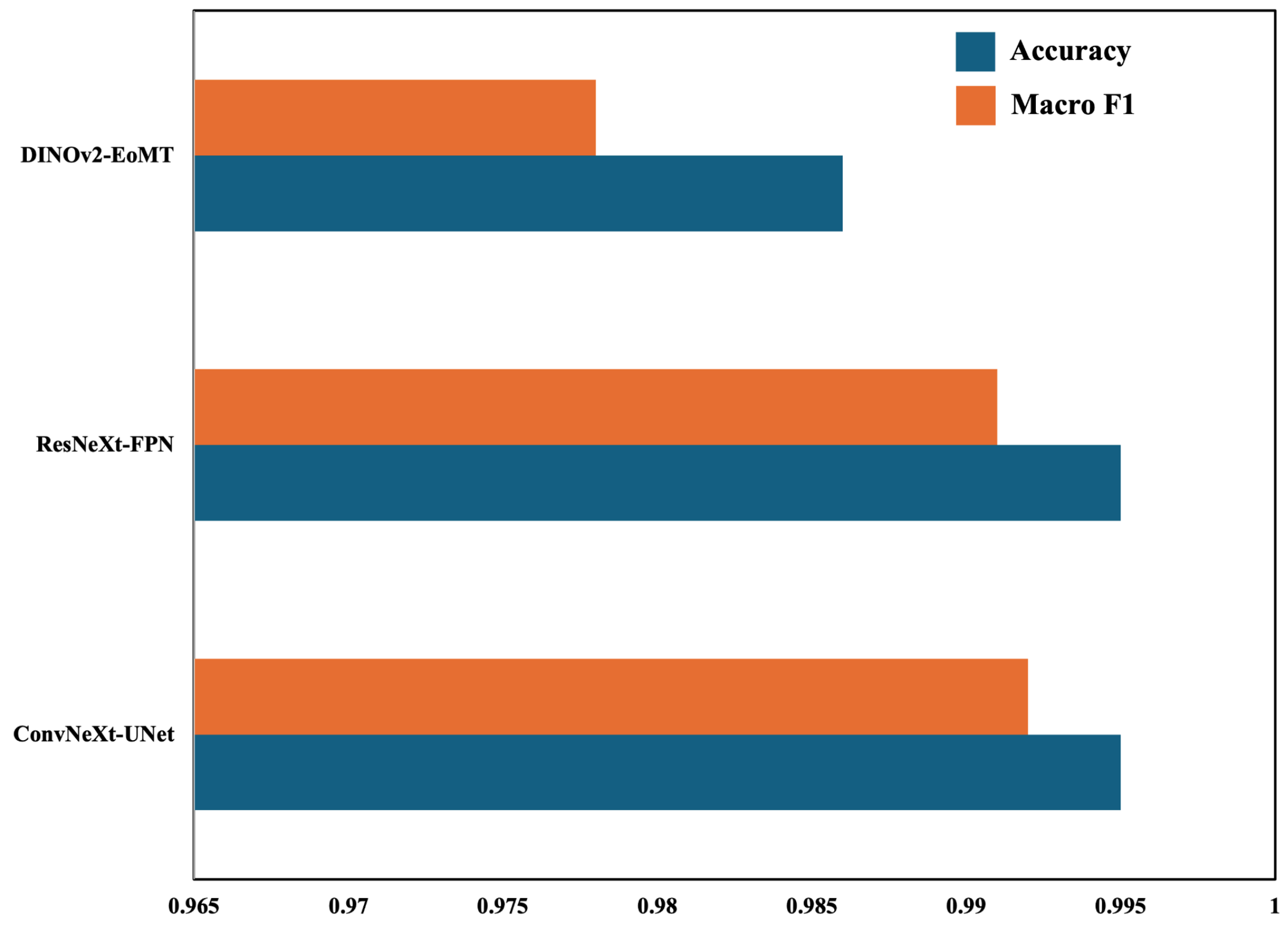

Figure 6: Comparison of accuracies of different model architectures trained with the same strategy. All the models displayed similar accuracies on the test data.

Importantly, the relatively small performance gap between the evaluated architectures suggests that the proposed mask preparation strategy and training pipeline are major contributors to the observed segmentation performance. This indicates that the framework is not tightly coupled to a specific network architecture and may be compatible with future segmentation backbones.

The predictions generated by the proposed framework are compared with the conventional manual histogram-based intensity thresholding approaches to segment a reconstructed slice from a basalt scan as shown in Fig 7. Traditional threshold-based methods showed substantial degradation in regions with overlapping intensity distributions, noise, and heterogeneous contrast conditions. In particular, porosity regions and low-contrast structural boundaries were often poorly separated using manual thresholding. Compared with manual thresholding, the proposed framework produced predictions that were less sensitive to noise and local intensity fluctuations. For this representative basalt slice, manual thresholding achieved a macro F1 score of 0.667, whereas the proposed ConvNeXt-UNet achieved a macro F1 score of 0.941, as shown in Table 2. The improvement was especially pronounced for porosity, where the model achieved an F1 score of 0.845 compared with 0.182 for manual thresholding. These results demonstrate that the deep learning approach is substantially more robust to the complex imaging conditions commonly encountered in synchrotron tomography experiments.

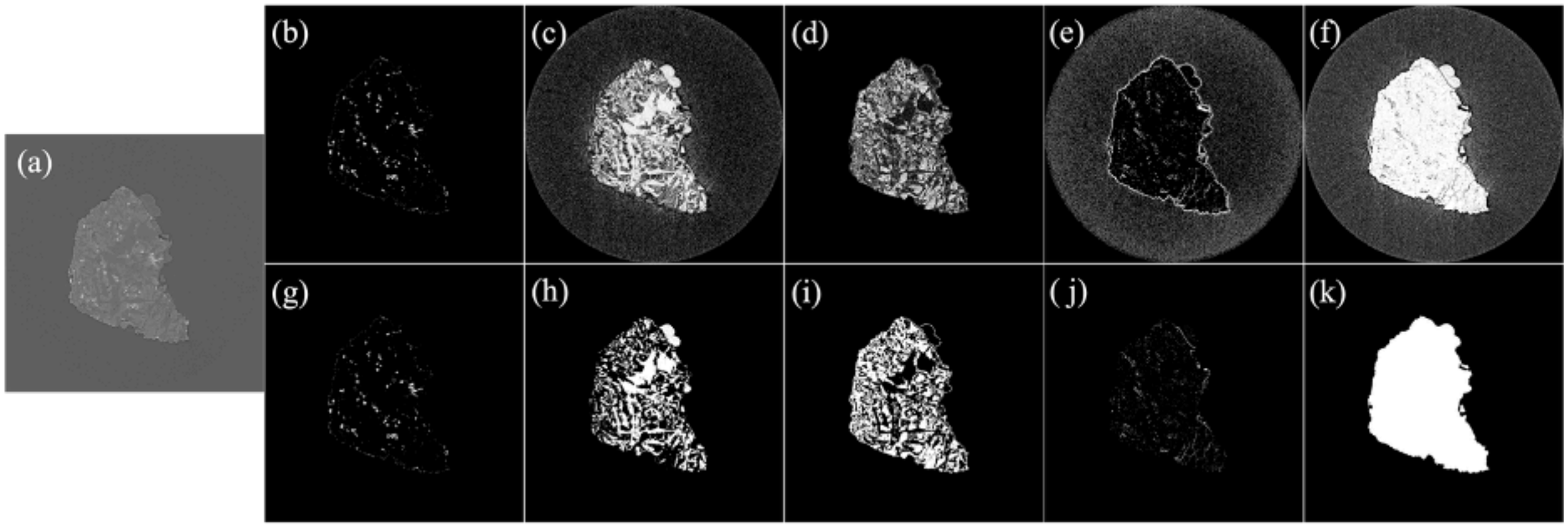

Figure 7: Comparison of model predictions with manual thresholding for all the classes. (a) reconstructed slice of basalt rock, (b - f) manual thresholding results of bright, dark gray, light gray, porosity and the sample, (g-k) model's predictions of the respective masks.

Table 2. Single-slice per-class segmentation performance comparison between manual thresholding and the proposed ConvNeXt-UNet model on a representative basalt slice.

| **Method** | **Accuracy** | **Macro F1** | **F1: Background** | **F1: Bright** | **F1: Dark Gray** | **F1: Light Gray** | **F1: Porosity** | **F1: Sample** |
|---|---|---|---|---|---|---|---|---|
| **Manual Thresholding** | 0.819 | 0.667 | 0.686 | 0.972 | 0.692 | 0.756 | 0.182 | 0.712 |
| **ConvNeXt-UNet** | 0.995 | 0.941 | 0.999 | 0.988 | 0.927 | 0.939 | 0.845 | 0.950 |

Overall, the results demonstrate that the proposed segmentation framework can provide rapid, stable, and physically meaningful segmentation of synchrotron tomography data across multiple material systems and imaging conditions. The framework achieves strong quantitative performance while maintaining stable qualitative performance on additional datasets with different microstructures and imaging conditions, making it well suited for real-time or near real-time deployment at synchrotron beamlines for diagnostic analysis and experimental decision-making.

# 4. Conclusion

In this work, we presented a zero-setup deployment framework for rapid multi-phase segmentation of synchrotron X-ray tomographic imaging data. The proposed approach combines a material-agnostic mask preparation strategy, multi-label semantic segmentation, class-aware sampling, and domain-robust preprocessing to produce diagnostic-level segmentations of previously unseen tomography datasets without requiring material-specific annotations, user prompting, or additional retraining. By decomposing reconstructed images into six broadly interpretable semantic regions (background, sample, bright regions, dark-gray regions, light-gray regions, and porosity), the framework provides an intuitive first-pass interpretation that can be generated within minutes of image reconstruction.

The framework was implemented using a ConvNeXt-UNet architecture and trained using only a small curated dataset of 25 annotated tomography images spanning multiple material classes. Despite the limited amount of labeled training data, the proposed approach consistently produced physically meaningful segmentations on previously unseen datasets and substantially outperformed conventional intensity-based thresholding, particularly in challenging regions containing overlapping intensity distributions, reconstruction artifacts, and porous microstructures.

The primary contribution of this work is the demonstration that practical, beamline-oriented segmentation can be achieved with zero user setup following model training. Rather than replacing application-specific segmentation models, the proposed framework provides an interpretable starting point that enables researchers to rapidly assess sample quality, identify regions of interest, detect potential experimental failures, and make informed decisions while experiments are still in progress. When higher segmentation accuracy or material-specific labeling is required, the generated masks can be refined manually or used to fine-tune specialized models.

More broadly, this work demonstrates the feasibility of integrating AI-assisted segmentation directly into synchrotron imaging workflows, helping bridge the gap between high-speed tomography reconstruction and actionable scientific interpretation. The proposed framework has the potential to substantially reduce the manual effort associated with tomography analysis, improve beamtime utilization through rapid experimental feedback, and provide a foundation for future scientific imaging systems based on foundation vision models and multimodal AI.

## 5. Acknowledgements

The authors thank Anyka Bergeson-Keller and Elly Shatsala of Lawrence Berkeley National Laboratory, Benite Ishimwe and Stuart McElhany of the University of California, Berkeley, and Qinxin Hu of the University of California, Santa Cruz, for contributing data used in this work. We also thank Dr. Xiaoya Chong, a research scientist from Photon Science Computing for reviewing and providing valuable feedback for this work.

This research used resources of the National Energy Research Scientific Computing Center (NERSC), a U.S. Department of Energy Office of Science User Facility located at Lawrence Berkeley National Laboratory. This research also used resources of the Advanced Light Source, a U.S. Department of Energy Office of Science User Facility operated under Contract No. DE-AC02-05CH11231. Pradyumna Elavarthi was supported in part by an Advanced Light Source Doctoral Fellowship in Residence. This work was additionally supported by the U.S. Department of Energy, Office of Science, Office of Basic Energy Sciences, Chemical Sciences, Geosciences, and Biosciences Division, through its Geosciences program at Lawrence Berkeley National Laboratory under Contract No. DE-AC02-05CH11231.

# 6. Data Availability

The datasets analysed during the current study are not publicly available as they are part of unpublished research but are available from the corresponding author on reasonable request.

# 7. Code availability

The underlying code for this study is available in Github and can be accessed via this link: https://github.com/pradyumnae/convnext-unet-segmentation

# Supplementary

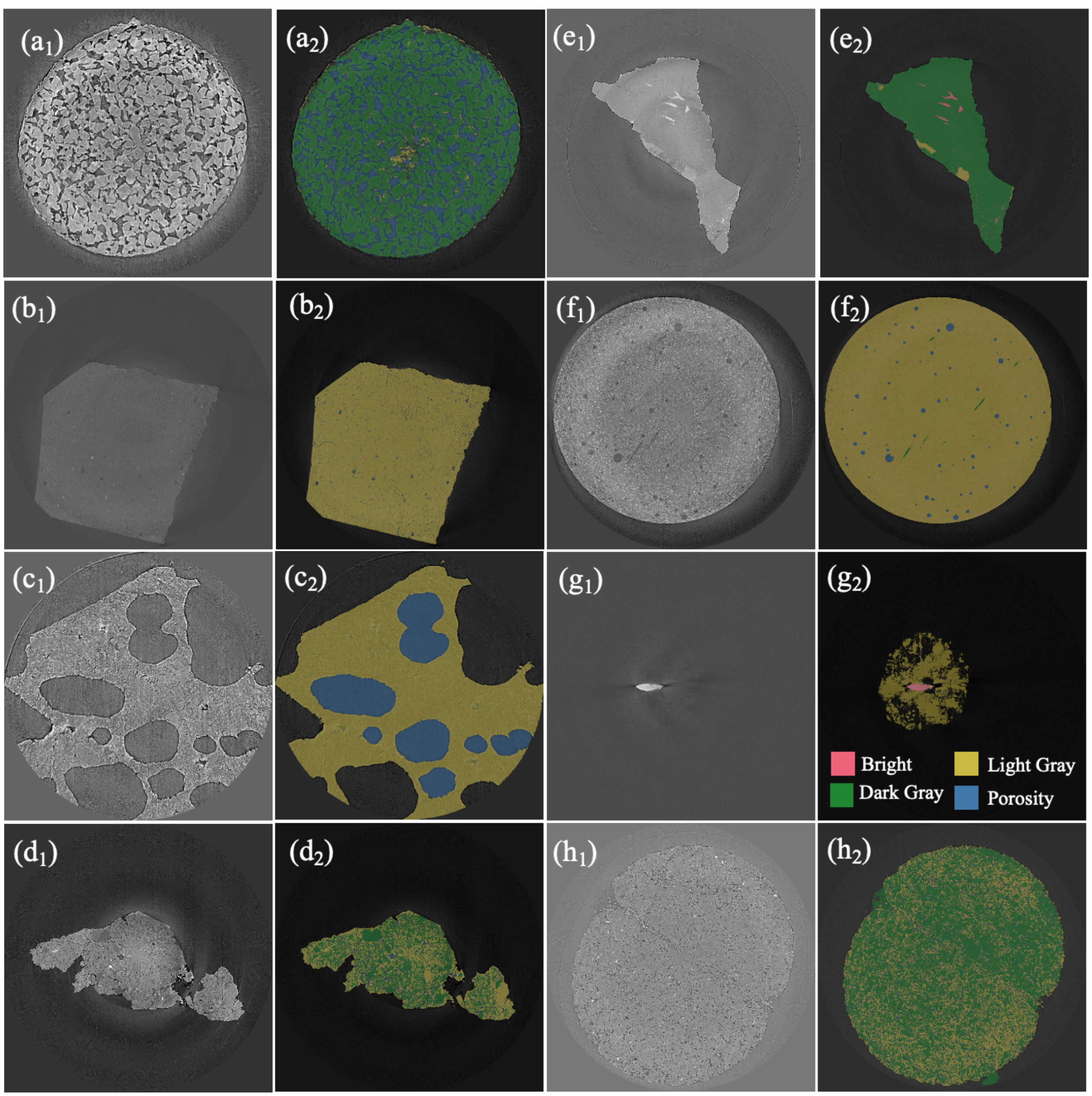

Supplementary Fig. S1. Original reconstructed images and their corresponding mask overlays for different micro CT samples. (a) Titanium PTL with green and yellow masks overlay solid Titanium particles and blue mask overlays pores. (b) Red clay and its light and dark gray phases. (c) Basaltic ash from East African virunga volcanic province. Yellow mask overlays ash particles, blue mask overlays pores. (d) Lunar regolith simulants. (e) Albite sample with light- and dark-gray attenuation phases, including possible fractures, pores, or mica-rich regions. (f)

Fiber-reinforced cement paste. Green mask overlays hydrated cement paste matrix, pink mask overlays unhydrated cement grains, blue masks overlay porosity in both capillary pores and entrapped air voids, and yellow mask overlays fibers which have been added to the cement paste to improve its strength. (g) Tantalum-rich metallic ore Coltan inside wooden sample holder. Pink mask overlays Coltan sample, yellow mask overlays wooden sample holder. (h) Sand sandwiched between rocks.